\documentclass[11pt,a4paper, table]{article}
\usepackage{enumitem}
\usepackage{float}
\usepackage{booktabs}
\usepackage{framed}
\usepackage{graphicx}
\usepackage{caption}
\usepackage[hyperref]{emnlp-ijcnlp-2019}
\usepackage{times}
\usepackage{latexsym}
\usepackage{subfigure}
\usepackage{url}
\usepackage{amsmath}
\usepackage{amssymb}
\usepackage{xspace}
\graphicspath{ {figures/} }
 \usepackage{multirow}
 \usepackage[normalem]{ulem}
 \useunder{\uline}{\ul}{}

\usepackage[ngerman]{babel}
\usepackage[utf8]{inputenc}
\usepackage[T1]{fontenc}

\usepackage{stackengine}

\aclfinalcopy 

\usepackage{tikz}
\usepackage{collcell}
\usepackage{etoolbox}

\newcommand*{\MinNumber}{0.0}%
\newcommand*{\MidNumber}{50.0} %
\newcommand*{\MaxNumber}{100.0}%

\newcommand{\gradientBLIMP}[1]{
    \ifdim #1 pt > \MidNumber pt
        \pgfmathparse{max(min(
        100.0*
        min(
        0.8*(#1 - \MidNumber)/((\MaxNumber-\MidNumber)/2), 
        0.5*(#1 - \MidNumber)/(\MaxNumber-\MidNumber)+0.5
        )
        ,100.0),0.00)} %
        \hspace{-0.33em}
        \xdef\tempa{\pgfmathresult}
        \cellcolor{black!35!green!\tempa!white!}{#1}
    \else
        \pgfmathparse{max(min(
        100.0*
        min(
        0.8*((\MaxNumber-\MidNumber) - #1)/((\MaxNumber-\MidNumber)/2),
        0.5*(\MidNumber - #1)/(\MidNumber-\MinNumber)+0.5
        )
        ,100.0),0.00)} %
        \hspace{-0.33em}
        \xdef\tempa{\pgfmathresult}
        \cellcolor{red!\tempa!white}{#1}
    \fi
 }

\newcommand*{\minval}{0.0}
\newcommand*{\maxval}{45.0}

\newcommand{\gradientBox}[1]{
    \ifdimcomp{#1pt}{>}{\maxval pt}{#1}{
    \ifdimcomp{#1pt}{<}{\minval pt}{#1}{
         \pgfmathparse{int(round(100*(#1/(\maxval-\minval))-(\minval*(100/(\maxval-\minval)))))}
        \xdef\tempa{\pgfmathresult}
        \cellcolor{black!10!blue!\tempa!white!} #1
    }}
 }
 
\newcommand{\mcrot}[4]{\multicolumn{#1}{#2}{\rlap{\rotatebox{#3}{#4}~}}} 

\newcommand\T{\rule{0pt}{3.1ex}}       
\newcommand\B{\rule[-1.7ex]{0pt}{0pt}} 

\title{Are word boundaries useful for unsupervised language learning?\thanks{\ To cite this work: Nguyen, T.A., de Seyssel, M., Algayres, R., Roze, P., Dunbar, E., Dupoux, E. (2020). Are word boundaries useful for unsupervised language learning? \textit{CoML Technical Report}, September 2020}}

\author{Tu Anh Nguyen$^{1,2}$, Maureen de Seyssel$^{1}$, Robin Algayres$^{1}$, Patricia Roze$^{1}$, \\ \bf Ewan Dunbar$^{1}$, Emmanuel Dupoux$^{1,2}$\\
$^1$ENS, INRIA, INSERM, UPEC, PSL Research University \\
$^2$Meta AI \\
\texttt{\{nguyentuanh208,emmanuel.dupoux\}@gmail.com} \\
}

\date{}

\begin{document}
\maketitle
\begin{abstract}
Word or word-fragment based Language Models (LM) are typically preferred over character-based ones in many downstream applications. This may not be surprising as words seem more linguistically relevant units than characters. Words provide at least two kinds of relevant information: boundary information and meaningful units. However, word boundary information may be absent or unreliable in the case of speech input (word boundaries are not marked explicitly in the speech stream). Here, we systematically compare LSTMs as a function of the input unit (character, phoneme, word, word part), with or without gold boundary information. We probe linguistic knowledge in the networks at the lexical, syntactic and semantic levels using three speech-adapted black box NLP psycholinguistically-inpired benchmarks (pWUGGY, pBLIMP, pSIMI). We find that the absence of boundaries costs between 2\% and 28\% in relative performance depending on the task. We show that gold boundaries can be replaced by automatically found ones obtained with an unsupervised segmentation algorithm, and that even modest segmentation performance gives a gain in performance on two of the three tasks compared to basic character/phone based models without boundary information.  


\end{abstract}

\section{Introduction}

Neural language models trained with a self-supervised objective have proven very successful as a pretraining method to learn useful representations. In particular, because they do not require labels, they can be trained on very large corpora taken from the internet, and then fine-tuned with a small amount of labels on downstream tasks \cite{elmo, devlin2018bert, radford2019languageGPT2, yang2019xlnet}. One of the unsolved problem is the optimality of the input units on which these neural models are trained \cite{sennrich-etal-2016-neural,bostrom2020}. Larger units like words tend to give better results, although they give rise to out-of-vocabulary (OOV) problems. Small units like characters do not have this  problem and may not require boundary information, but give rise to slightly lower performance. Word part units like BPE \cite{gage1994,sennrich-etal-2016-neural} seem to be a good compromise, providing larger units but allowing to deal with unseen words. Note that BPEs require word boundaries, even if they are modeling subword parts.  

Recent work has applied self-supervised Language Modeling (LM) or masking objectives to raw audio, totally by-passing text, basing the loss function on automatically discovered quantized speech units \cite{baevski2020wav2vec,baevski2019effectiveness}. Even though this approach has been shown to be very useful for pretraining an ASR system with few labels, the question remains as to what would be the optimal kinds of units for language modeling from raw speech. This may become even more salient, as quantized speech units tend to be smaller than phonemes (therefore unlikely units to carry meaning or syntactic information), and without any word boundary (making it difficult to define meaningful higher order units). In other words, if we want to apply LM approaches to raw audio, a major stumbling block may be the word segmentation problem. The fact that word segmentation from audio is itself a difficult problem may give rise to a circularity issue: we may need accurate word segmentation in order to do proper language modeling from audio without any labels. We may need excellent acoustic units to do accurate word segmentation. We may need very good language modeling in order to obtain accurate decoding into acoustic units. Back to square one.

Here, we wish to estimate, as a preliminary question, the cost of switching from a word-based representation (with boundaries) to a phoneme one, without boundaries. We  use the Librispeech  corpus \cite{Panayotov2015librispeech}, for which we have both the text transcription and a phoneme-based transcription. We use use phoneme transcriptions as a proxy for 'accurate' acoustic units, leaving for later the problem of erroneous transcripts when the units are derived from speech. We also test the possibility of replacing gold word boundaries by automatically obtained one using an unsupervised word segmentation algorithm.  

When comparing LMs with widely different kinds of input units, standard metrics like perplexity cannot be used because these metrics scale in complicated ways with the granularity of the input units. Instead here, we rely on three psycholinguistically inspired black-box NLP benchmarks which are independent of unit granularity, and which we adapt to be speech-compatible by phonemizing them and filtering the vocabulary with the Librispeech train set. The first two are based on assigning pseudo-probabilities to input strings, which are used them as a proxy for an acceptability score. For the lexical benchmark (pWUGGY), we compare the acceptability of words (like ``brick") to that of a non-word (like ``blick"). The words and non-words are otherwise matched on unigram and bigram probabilities. For the syntactic benchmark (pBLIMP), we adapted and phonetically transcribed the BLIMP dataset \cite{warstadt2019blimp} in which the acceptability of pairs of grammatical and ungrammatical sentences is assessed. The semantic test (pSIMI) is based on the distance between embeddings of words, which is correlated with human obtained distances. 

The structure of the paper is as follows: after presenting the related work (Section \ref{sec:related}) and methods (datasets, models and metrics, Section \ref{sec:method}), we present the results of baseline character-based LSTM models with access to word boundaries (Section \ref{sec:baselines}). We then present experiments where we change the units to be phones, and remove the gold boundaries, or replace them with automatically extracted ones (Section \ref{sec:nobound}). 

\vspace{-1em}
\section{Related work}\label{sec:related}
\vspace{-.5em}

\textbf{Units for LSTMs}
The importance of word boundaries has been investigated by \citet{hahn2019}. They compared word based and character based LSTMs where the word boundaries (space character) were removed, on a variety of probe tasks. They found that the character-based LSTMs passed a number of linguistic tests, sometimes better than word based models that are impaired by the presence of OOVs. Here, we follow the same inspiration, but evaluate more systematically models that are boundary based, but do not suffer from OOVs (ie, BPE and fallback models), in order to give word models a fairer comparison point and provide a quantitative measure of the cost of not having boundaries. We also expand the investigation to phoneme representations that are step closer to speech.

\textbf{Black box linguistics}
Among the variety of Black-Box linguistic tasks, psycholinguistically inspired ones enable the direct comparison of models and humans.  Grammaticality judgments for recurrent networks have been investigated since \citet{allen1999:emergence}, who use closely matched pairs of sentences to investigate grammatical correctness. This approach has been adopted recently to assess the abilities of RNNs, and LSTMs in particular, to capture syntactic structures. For instance, \citet{linzen:2016} and \citet{gulordava:2018} use word probes in minimally different pairs of English sentences to study number agreement. To discriminate grammatical sentences from ungrammatical ones, they retrieve the probabilities of the possible morphological forms of a target word, given the probability of the previous words in the sentence. Practically, in the sentence ``the boy \underline{is} sleeping'', the network has detected number agreement if \textit{$\mathbf{P}(w = is) > \mathbf{P}(w  = are)$}. This methodology has also been adapted by \citet{goldberg:2019} to models trained with a masked language-modeling objective. Those works find that in the absence of many detractors or complex sentence features, recent language models perform well at the number-agreement problem in English.

More closely related to our work, \citet{ravfogel2018can} use word probes to examine whether LSTMs understand Basque agreement. Like German, Basque is a morpho-syntactically rich language with relatively free word order, thus providing a challenging setting for the LM. In contrast to our work, the LM's ability to understand verb argument structure is tested on number-agreement and on suffix recovery tasks, which involve localized changes rather than whole sentence perturbations and re-orderings.

\begin{table*}[]
\centering
\begin{tabular}{lll}
\hline
\vspace{-1em}\\
\bf{dataset}&\bf{sub-dataset}&\bf{examples}   \\
\hline
\vspace{-1em}\\
 &  & Heading - Heasing       \\
\multirow{-2}{*}{\parbox{2cm}{pWUGGY}}& \multirow{-2}{*}{-}  & Squalled - Squilled \\
\hline
\vspace{-1em}\\
        &  anaphor gender agreement   & Katherine can’t help \textit{herself}.\\
        &     & Katherine can’t help \textit{himself}.       \\
        &  irregular p.participle adj.   & The \textit{forgotten} newspaper article was bad.      \\
\multirow{-3}{*}{\parbox{2cm}{pBLIMP}}&  & The \textit{forgot} newspaper article was bad. \\
\hline
\vspace{-1em}\\
        & MEN & (Abandoned , Ruins) - 6.4      \\
        & MEN & (Abstract, Frog) - 0.8 \\
        & simverb-3500 & (Abduct, Kidnap) - 8.63 \\
\multirow{-3}{*}{\parbox{2cm}{pSIMI}} & simverb-3500 & (Abduct, Tap) - 0.5   \\
\hline
\end{tabular}
\caption{Example of tests tokens from the three benchmarks as described in section \ref{sec:tests}}
\end{table*}

\vspace{-1em}
\section{Methods}\label{sec:method}
\vspace{-.5em}

\subsection{Training set} We used as a training set the transcription of the Librispeech 960h dataset \cite{Panayotov2015librispeech}, composed of 281K sentences (9M words, 40M characters or 33M phonemes). We can therefore give a comparative performance with other speech-based work. As it is the transcription of an ASR dataset, the text has originally been cleaned, removed all the punctuation marks and uppercased, resulting in a vocab size of 90K. For the phonetic transcription, we used the original LibriSpeech lexicon, for some words that are not in the lexicon, we used the G2P-seq2seq toolkit \footnote{https://github.com/cmusphinx/g2p-seq2seq} to generate their phonetic transcriptions.


\subsection{Black Box test sets}
\label{sec:tests}

We setup three tasks, to evaluate the LMs at three levels: the lexicon (the pWUGGY benchmark), syntax (the pBLIMP benchmark) and semantics (the pSIMI benchmark). All of these benchmarks are presented in two formats: a character format (in which case the test tokens are in text) and a phonetic format, obtained by using the same G2P-seq2seq toolkit as for the train set.

\paragraph{Lexicon - the pWUGGY benchmark.} We built on \citet{godais2017charnlm} which used the 'Spot-the-word' task in which the networks are presented with a pair of an existing word and a matching non-word, and are evaluated on their capacity to attribute a higher probability to the word.

The non-words are generated with the WUGGY software \cite{keuleers2010wuggy}, which generates for a given word, a list of candidate nonwords matched in phonotactics, syllabic structure, and other character-based constraints of the English language. We added additional constraints using a stochastic sampler to also match unigram and bigram, character and phoneme frequencies (see Supplementary Material \ref{SM:sampling} for more details).


The test dataset is composed of two subsets: a set of pairs built with words present in LibriSpeech training set and a set of pairs built with words not existing in LibriSpeech (OOV words) with 30K and 10K pairs respectively. We also prepared a small development set containing 10K pairs of words from LibriSpeech disjoint from the test set in case of necessary uses.
Each word or nonword in a pair was then preceded and followed by a \texttt{<EOS>} symbol to help the model distinguish a word from a prefix or suffix (e.g., a nonword \textit{firew} and a word \textit{firework}). 



\paragraph{Sentence Grammaticality - the pBLIMP benchmark.}  

This Benchmark is adapted from BLIMP \cite{warstadt2019blimp}, a dataset of linguistic minimal sentence pairs of matched grammatical and ungrammatical sentences. As for the preceding test, the task is to decide which of the two members of the pair is grammatical or not based on the probability of the sentence. 

We adapted the code used to generate the BLIMP dataset \cite{warstadt2019blimp} in order to create pBLIMP, specifically tailored for speech purposes. In BLIMP, sentences are divided into twelve broad categories  focusing  on  different linguistic paradigms in the fields of syntax, morphology or semantics. These categories are themselves divided into 67 finer linguistic subcategories, containing 1000 sentence pairs each, automatically generated using expert hand-crafted grammar. One additional subcategory was also subsequently added in the code.

To make this dataset 'speech-ready', we discarded five subcategories and slightly modified the grammar for 9 additional subcategories in order to avoid any difficulty of generating a prosodic contour for the ungrammatical sentences. We also removed from the vocabulary all words not present in the Librispeech \cite{Panayotov2015librispeech} train set, as well as compound words and homophones that could cause further understanding issues once synthesised. 5000 sentence pairs were then generated for each of the 63 remaining subcategories. 
We then sampled sentence pairs from the generated pool to create a development and a test set, ensuring that the larger linguistic categories were sampled in terms of n-gram language model scores (see Supplementary Material \ref{SM:sampling}). The test and development sets contains 63000 and 6300 sentence pairs respectively, with no sentence pairs overlap.




\paragraph{Semantics: the pSIMI benchmark.} Here, the task is to compute the similarity of the representation of pairs of words and compare it to human similarity judgements. 

Based on previous work \citet{chung2018speech2vec}, we used a set of 13 existing semantic similarity/relatedness tests.  
The similarity-based datasets include WordSim-353 \cite{yang2006verb},
WordSim-353-SIM \cite{agirre2009study}, mc-30 \cite{miller1991contextual}, rg-65 \cite{rubenstein1965contextual}, Rare-Word (or rw) \cite{luong2013better}, simLex999 \cite{hill2015simlex},
simverb-3500 \cite{gerz2016simverb}, verb-143 \cite{baker2014unsupervised} , YP-130 \cite{yang2006verb}
and the relatedness-based datasets include MEN \cite{bruni2012distributional}, Wordsim-353-REL \cite{agirre2009study}, mturk-287 \cite{radinsky2011word}, mturk-771 \cite{halawi2012large}.

All scores were normalised on a 0-10 scale, and pairs within a same dataset containing the same words in different order were averaged.  Pairs containing a word absent of the LibriSpeech train set \cite{Panayotov2015librispeech} were discarded.
We  selected as development set the mturk-771 dataset, which was, in a preliminary study using character and word-based LMs, both highly correlated with all other datasets and was large enough to be used as a development set. It was also ensured that no pair from the development set was present in any of the test sets. All other 12 datasets were used as test sets.



\subsection{Automatic segmentation of sentences}
We used an unsupervised word segmentation method called DP-Parse, which is inspired from the the work of \citet{goldwater}, which had slightly lower segmentation scores, but a 50 times speedup (see Supplementary Material \ref{SM:DPPARSE} for more details), and train it on the unsegmented version of the training dataset on phoneme and character levels. The word segmentation results are shown in table \ref{tab:WordsegResults}.
\begin{table}[h]
    \centering
    \resizebox{0.5\textwidth}{!}{
    \begin{tabular}{|c|c|c|c|c|c|c|}
    \hline
        \multirow{2}{*}{level} & \multicolumn{3}{c|}{token scores} & \multicolumn{3}{c|}{boundary scores} \\
        \cline{2-7}
         & f-score & precision & recall & f-score & precision & recall \\
         \hline
        char & 43.57 & 48.19 & 39.75 & 75.07 & 83.34 & 68.30 \\
        \hline
        phone & 46.37 & 51.18 & 42.38 & 77.09 & 85.39 & 70.25 \\
        \hline
    \end{tabular}}
    \caption{Automatic word segmentation scores on the transcription of LibriSpeech-960 dataset}
    \label{tab:WordsegResults}
\end{table}

After training the word segmentation models, we parsed the unsegmented training dataset and obtained two new corpus with charseq and phoneseq unit level respectively, which is similar to word unit level in the segmented dataset. We also used the trained models to parse all the test sets into charseq and phoneseq unit level respectively.

\subsection{Language Models}

\begin{figure*}[]
\centering
{\includegraphics[width=0.98\textwidth]{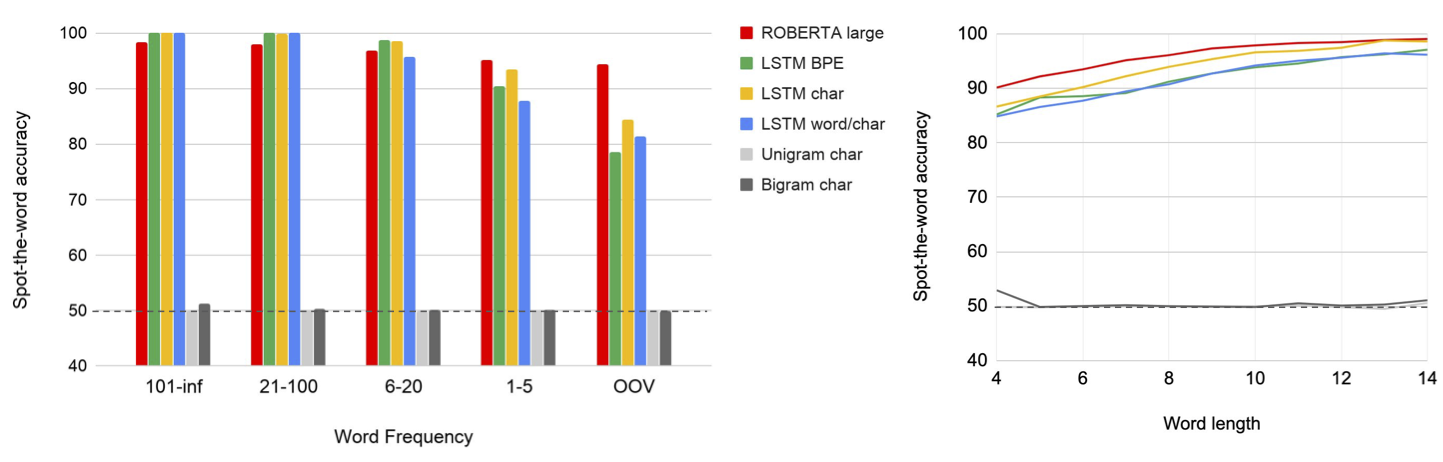}}
\caption{{\bf Spot-the-word accuracy} (pWUGGY test set, higher is better, chance level at 50\%) for LSTM models trained on the transcription of LibriSpeech-960 corpus on different types of units (character, BPE, word), as a function of word frequency (left) and word length (right). OOV corresponds to words unseen in the training set. For comparison, the performance of a ROBERTA Large pretrained model and character unigram and bigram baselines.}
\label{fig:WUGGY}
\end{figure*}

\label{sec:models}
We used classic LSTM models \cite{hochreiter1997long} for language modeling. We introduced several training modes, according to the units used to train the model. In the char (phon) model, we used the character (or phonetic) transcription to train the model. By default, we leave a special \texttt{<SPACE>} character between words, unless specified otherwise. For subword-unit models, we used a slightly modified version of Byte-Pair Encoding (BPE) \cite{sennrich-etal-2016-neural} that can handle both character and phonetic units. In the BPE models, 
we define 20K BPE units on the training set. In the word/char model, we establish a lexicon, which we truncate at 20K and then replace the oovs with their character transcriptions.

Following \citet{hahn2019}, we used a three-layer LSTM with an embedding layer of 200 units and two hidden layers of 1024 units for character and phoneme-level models. For word and subword models, we used a two-layer LSTM with hidden (and embedding) size of 1024 units. During training, each sentence is presented preceded and followed by a \texttt{<EOS>} symbol.
The training was done using the fairseq library \cite{ott-etal-2019-fairseq}, we evaluate the models on the corresponding valid-clean set of LibriSpeech and take the best model based on evaluation loss. 

For top-line comparison, we used a pretrained ROBERTA model \cite{Liu2019ROBERTA}, which is a 24-layer Transformer \cite{dai2018transformer} trained with a masked language model objective on 50K BPE subword units on a huge dataset of total 160GB, 3000 times bigger than our LibriSpeech transcription dataset.
\vspace{-0.6em}
\subsection{Language Model Pseudo-probability Scores}
\label{sec:scoring}
Each LSTM language model can assign a probability score for a sequence of tokens $s_1...s_N$ by using the decomposition of the joint probability
\vspace{-0.6em}
\begin{equation*}
    P(s_{1}..s_{N})=\prod_{i=1}^N P(s_{i}|s_{1}..s_{i-1}),
\end{equation*}
\vspace{-0.em}
where each conditional probability $P(s_{i}|s_{1}..s_{i-1})$ is estimated by the softmax output of the symbol $s_{i}$ given its preceding context $s_{1},\dots,s_{i-1}$ fed as input to the LSTM. 

For the ROBERTA model, we used a \emph{pseudo probability} (PP) score \cite{salazar-etal-2020-masked} obtained by multiplying the conditional probability of each token $s_{i}$ given all the other tokens 
\vspace{-0.6em}
\begin{equation*}
    PP(s_{1}..s_{N})=\prod_{i=1}^N P(s_{i}|s_{1}..s_{i-1}s_{i+1}..s_{N}),
\end{equation*}
where $P(s_{i}|s_{1}..s_{i-1}s_{i+1}..s_{N})$ is estimated as the softmax output of the token $s_{i}$ given the input $s_{1}..s_{i-1}\texttt{<mask>}s_{i+1}..s_{N}$ to the ROBERTA model.

The spot-the-word accuracy for the pWUGGY test is then computed as the average of the indicator function $1_{score(word_k)>score(nonword_k)}$ over the test set of pairs ${(word_k,nonword_k)}$. Similarly with grammatical and ungrammatical pairs of sentences for the pBLIMP test.

\subsection{Language Model Distance Scores}
\label{sec:distanceScoring}
Neural language models can compute a fixed-length representation vector for each sequence of tokens $s_1...s_N$ by simply aggregating the outputs of a hidden layer with a pooling function
\begin{equation*}
    v(s_1...s_N) = f_{pool}\left(h^{(i)}_1...h^{(i)}_N\right),
\end{equation*}
where $f_{pool}$ is the pooling function and $h^{(i)}_1,...,h^{(i)}_N$ are the outputs of the $i^{th}$ hidden layer of the network.
The distance score of two sequences of tokens $s_1...s_N$ and $t_1...t_M$ is then computed as the cosine similarity between the representation vectors of the two sequences.

We then compute the semantic similarity score as the Spearman’s rank correlation coefficient $\rho$ between the distance scores given by the model and the true human scores in the pSIMI test. 

It's worth noting that the choice of the pooling function $f_{pool}$ as well as the hidden level $i$ can greatly affect the similarity scores and thus need to be optimised. Therefore, for each model, we choose the pooling function and the hidden level that gives the best score on the dev set, and report the corresponding score on the test set.

\begin{table*}[h!]
    \begin{center}
    \resizebox{\textwidth}{!}{
        \addtolength\tabcolsep{2pt}
        \begin{tabular}{r*{13}{c}}
          & \mcrot{1}{c}{45}{\textbf{Overall}}    & \mcrot{1}{c}{45}{Ana. Agr.}    & \mcrot{1}{c}{45}{Agr. Str.}    & \mcrot{1}{c}{45}{Binding}    & \mcrot{1}{c}{45}{Ctrl. Rais.}    & \mcrot{1}{c}{45}{D-N Agr.}    & \mcrot{1}{c}{45}{Ellipsis}    & \mcrot{1}{c}{45}{Fill. Gap.}    & \mcrot{1}{c}{45}{Irregular}    & \mcrot{1}{c}{45}{Island} & \mcrot{1}{c}{45}{NPI Li.} & \mcrot{1}{c}{45}{Quantifiers} & \mcrot{1}{c}{45}{S-V Arg.} \\
          \T unigram char \ \ \B &  \textbf{\gradientBLIMP{44.01}} & \gradientBLIMP{50.53} & 49.3 & \gradientBLIMP{52.06} & \gradientBLIMP{38.04} & \gradientBLIMP{49.61} & \gradientBLIMP{52.55} & \gradientBLIMP{49.59} & \gradientBLIMP{47.34} & \gradientBLIMP{37.19} & \gradientBLIMP{28.31} & \gradientBLIMP{40.07} & \gradientBLIMP{42.96} \\
          \T unigram word \ \ \B & \textbf{\gradientBLIMP{47.96}} & 50.8 & \gradientBLIMP{49.99} & \gradientBLIMP{65.04} & \gradientBLIMP{37.06} & \gradientBLIMP{51.81} & \gradientBLIMP{51.49} & \gradientBLIMP{87.43} & \gradientBLIMP{14.89} & \gradientBLIMP{20.62} & 29.1 & \gradientBLIMP{47.27} & \gradientBLIMP{45.69} \\
          \T bigram char \ \ \B & \textbf{\gradientBLIMP{48.19}} & \gradientBLIMP{50.59} & \gradientBLIMP{48.78} & \gradientBLIMP{50.43} & \gradientBLIMP{42.79} & \gradientBLIMP{49.68} & \gradientBLIMP{52.55} & \gradientBLIMP{73.64} & \gradientBLIMP{14.89} & 41 & \gradientBLIMP{28.74} & \gradientBLIMP{50.35} & \gradientBLIMP{53.55} \\
          \T bigram word \ \ \B & \textbf{\gradientBLIMP{52.05}} & \gradientBLIMP{49.41} & \gradientBLIMP{50.04} & \gradientBLIMP{70.64} & \gradientBLIMP{42.43} & \gradientBLIMP{52.54} & \gradientBLIMP{49.36} & \gradientBLIMP{60.07} & \gradientBLIMP{16.06} & \gradientBLIMP{47.78} & \gradientBLIMP{43.16} & \gradientBLIMP{66.99} & \gradientBLIMP{57.43} \\
          \T LSTM char  \ \ \B & \textbf{\gradientBLIMP{63.02}} & \gradientBLIMP{67.50} & \gradientBLIMP{56.21} & \gradientBLIMP{79.94} & \gradientBLIMP{58.53} & \gradientBLIMP{82.93} & \gradientBLIMP{81.70} & \gradientBLIMP{75.17} & \gradientBLIMP{22.02} & \gradientBLIMP{53.43} & \gradientBLIMP{43.72} & \gradientBLIMP{71.81} & \gradientBLIMP{58.53} \\
          \T LSTM word BPE \ \ \B & \textbf{\gradientBLIMP{65.92}} & \gradientBLIMP{86.35} & \gradientBLIMP{58.71} & \gradientBLIMP{80.00} & \gradientBLIMP{61.06} & \gradientBLIMP{84.74} & \gradientBLIMP{63.30} & \gradientBLIMP{61.61} & \gradientBLIMP{86.25} & \gradientBLIMP{54.81} & \gradientBLIMP{46.19} & \gradientBLIMP{85.47} & \gradientBLIMP{62.73} \\
          \T LSTM word \ \ \B & \textbf{\gradientBLIMP{66.68}} & \gradientBLIMP{87.90} & \gradientBLIMP{59.84} & \gradientBLIMP{78.16} & \gradientBLIMP{60.38} & \gradientBLIMP{77.88} & \gradientBLIMP{69.00} & \gradientBLIMP{61.49} & \gradientBLIMP{83.70} & \gradientBLIMP{55.91} & \gradientBLIMP{56.71} & \gradientBLIMP{92.77} & \gradientBLIMP{62.67} \\
          \T ROBERTA large \ \ \B & \textbf{\gradientBLIMP{82.06}} & \gradientBLIMP{97.70} & \gradientBLIMP{75.68} & \gradientBLIMP{82.32} & \gradientBLIMP{79.74} & \gradientBLIMP{95.69} & \gradientBLIMP{93.30} & \gradientBLIMP{73.90} & \gradientBLIMP{88.00} & \gradientBLIMP{69.17} & \gradientBLIMP{81.43} & \gradientBLIMP{92.73} & \gradientBLIMP{87.42} 
        \end{tabular}}
    \end{center}
    \caption{{\bf Sentence acceptability accuracy} (pBLIMP test set, higher is better, chance level at 50\%) for LSTM models trained on the transcription of LibriSpeech-960 on different types of units (character, BPE, word), as a function of syntactic phenomenon. For comparison, the performance of a pretrained ROBERTA large model and two baseline character and word bigram models.}
    \label{table:BLIMP}
\vspace{-1.5em}
\end{table*}

\section{Baseline results}
\vspace{-0.7em}
\label{sec:baselines}
\subsection{Lexicon baselines}
\vspace{-0.5em}

In Figure \ref{fig:WUGGY}, we present the results of the pWUGGY test set on three LSTM language models trained on the librispeech dataset: character-based LSTM, BPE-based LSTM, and word/char LSTM. The last one contains a 20k word lexicon with fallback on character. We also present the results of a ROBERTA model based on BPE. 

The spot-the-word accuracy (Figure \ref{fig:WUGGY}, left) is overall very high for all models (>90\%) and shows a frequency effect (lower frequency being less accurate than high frequency). The models show better than chance performance for OOVs (around 80\%). For the three LSTMs, this suggests that they are able to generalize lexicality beyond the words in the training set, presumably through morphological generalizations. For ROBERTA of course,  the training set was much larger, and many of our pWUGGY OOVs may have been seen. We did not include the results of a word LSTM, since such a system replaces all of the nonwords, as well as many unfrequent words not in its lexicon, with the symbol <UNK>, yielding actually an average score below chance (since the probability of <UNK> turns out to be higher than many test words) -- the result is shown in Table \ref{tab:res} below.  Also note that the unigram and bigram models are very close to the chance level, which attests to the fact that the pWUGGY dataset was indeed well matched on unigram and bigram frequency.

In the right of the Figure \ref{fig:WUGGY}, we observe that generally the spot-the-word accuracy increases with the length of the words, which may be due to the fact that the phonetic space is sparser for long than for short words. As a consequence, a short nonword like "tup" could be continued as a real word in multiple wats ("tuple", "tupperware", etc.), which means that the distinction between words and nonwords comes towards the end of the string. In contrast, a long nonword can rarely be salvaged into a word (eg, 'rhanoceros' is a nonword very early on). 

\begin{table*}[]
    \begin{center}
    \resizebox{\textwidth}{!}{
    \setlength\tabcolsep{10pt}
    \stackunder[10pt]{\begin{tabular}{r|*{25}{c}}
      layer \ \ & 1 & 2 & 3 \\
      \hline
      \T mean \ \B & \gradientBox{1.79} & \gradientBox{0.94} & \gradientBox{0.25} \\
      \T max \ \B & \gradientBox{3.48} & \gradientBox{3.23} & \gradientBox{2.82} \\
      \T min \ \B & \gradientBox{5.75} & \gradientBox{0.83} & \gradientBox{1.44} \\
    \end{tabular}}{LSTM char gold bound} \quad \quad \quad 
    
    
    \stackunder[10pt]{\begin{tabular}{r|*{25}{c}}
      layer \ \ & 0 & 1 & 2 \\
      \hline
      \T mean \ \B & \gradientBox{15.14} & \gradientBox{16.19} & \gradientBox{15.23} \\
	  \T max \ \B & \gradientBox{17.34} & \gradientBox{16.71} & \gradientBox{13.80} \\
	  \T min \ \B & \gradientBox{17.80} & \gradientBox{16.81} & \gradientBox{14.53} \\
    \end{tabular}}{LSTM BPE} \quad \quad \quad 
    
    \stackunder[10pt]{\begin{tabular}{r|*{25}{c}}
      layer \ \ & 0 & 1 & 2 \\
      \hline
      \T mean \ \B & \gradientBox{22.84} & \gradientBox{27.90} & \gradientBox{26.00} \\
	  \T max \ \B & \gradientBox{24.39} & \gradientBox{26.70} & \gradientBox{23.31} \\
	  \T min \ \B & \gradientBox{20.29} & \gradientBox{26.37} & \gradientBox{23.75} \\
    \end{tabular}}{LSTM word} }\\
    \vspace{0.5em}
    \resizebox{\textwidth}{!}{
        \setlength\tabcolsep{0pt}
        \stackunder[10pt]{\begin{tabular}{r|*{25}{c}}
          layer \ \ & 0 & 1 & 2 & 3 & 4 & 5 & 6 & 7 & 8 & 9 & 10 & 11 & 12 & 13 & 14 & 15 & 16 & 17 & 18 & 19 & 20 & 21 & 22 & 23 & 24 \\
          \hline
          \T mean \ \B & \gradientBox{11.38} & \gradientBox{10.46} & \gradientBox{8.82} & \gradientBox{9.96} & \gradientBox{12.87} & \gradientBox{15.52} & \gradientBox{19.76} & \gradientBox{23.94} & \gradientBox{25.93} & \gradientBox{24.18} & \gradientBox{24.72} & \gradientBox{24.79} & \gradientBox{22.49} & \gradientBox{22.20} & \gradientBox{21.82} & \gradientBox{20.59} & \gradientBox{17.92} & \gradientBox{17.78} & \gradientBox{18.06} & \gradientBox{19.17} & \gradientBox{19.87} & \gradientBox{18.16} & \gradientBox{18.71} & \gradientBox{17.14} & \gradientBox{7.20} \\
          \T max \ \B & \gradientBox{12.02} & \gradientBox{13.31} & \gradientBox{7.58} & \gradientBox{10.47} & \gradientBox{14.54} & \gradientBox{15.88} & \gradientBox{18.65} & \gradientBox{24.31} & \gradientBox{30.46} & \gradientBox{30.86} & \gradientBox{29.73} & \gradientBox{29.08} & \gradientBox{25.21} & \gradientBox{23.91} & \gradientBox{23.95} & \gradientBox{23.52} & \gradientBox{18.87} & \gradientBox{17.95} & \gradientBox{18.72} & \gradientBox{19.30} & \gradientBox{20.76} & \gradientBox{19.73} & \gradientBox{17.62} & \gradientBox{15.71} & \gradientBox{5.15} \\
          \T min \ \B & \gradientBox{10.57} & \gradientBox{11.47} & \gradientBox{9.60} & \gradientBox{11.67} & \gradientBox{15.18} & \gradientBox{19.98} & \gradientBox{23.53} & \gradientBox{28.12} & \gradientBox{32.28} & \gradientBox{29.93} & \gradientBox{29.81} & \gradientBox{30.97} & \gradientBox{27.45} & \gradientBox{26.27} & \gradientBox{26.01} & \gradientBox{24.90} & \gradientBox{21.97} & \gradientBox{20.98} & \gradientBox{21.03} & \gradientBox{22.33} & \gradientBox{23.38} & \gradientBox{18.55} & \gradientBox{18.64} & \gradientBox{17.27} & \gradientBox{5.38} \\
        \end{tabular}}{ROBERTA large} }
    \end{center}
    \caption{\textbf{Semantic similarity scores} (Spearman's correlation with human judgement, higher is better) on the pSIMI development set for 3 LSTM models with different types of units (character, BPE, word) and a ROBERTA large model as a function of the hidden level of the representation outputs and the pooling functions.}
    \label{table:SEMANTICS}
\vspace{-1.3em}
\end{table*}

\vspace{-0.7em}
\subsection{Syntactic judgments baselines}
\vspace{-0.5em}

Here, we present the results of pBLIMP on the same three LSTM models and baselines discussed above (see Table \ref{table:BLIMP}).

First, our introduction of unigram and bigram character and word controls show that even though the pBLIMP dataset is overall well matched, and despite our attempt to reequilibrate this dataset, 
certains paradigms are not. Most notably Binding datset and Filler-Gap datasets can have a significant above chance score in word baseline model. Vice versal the Irregular and NPI datasets are below chance in baseline models. It is useful to have these simple baselines in order not to over-interpret the results of more complicated language models. In this respect for instance, no model can be claimed to really solve the filler Gap dataset, since no model beats the unigram word model. 

Second, and unsurprisingly, ROBERTA is able to outperform all models and corresponds to a topline in our case. It beats all of the simple baselines (except the Filler-Gap dataset). 

Third, BPE and word models are on average better than character LMs, although in some categories this is the other way around. The difference between the three classes of LSTMs is way smaller than the difference between any of these models and ROBERTA. This is compatible with Baroni's claim that character LMs are almost as good as models based on higher order units.

\vspace{-0.7em}

\subsection{Semantic similarity baselines}
\vspace{-0.5em}

Here, we present the results of pSIMI task on the same three LSTM models plus ROBERTA. As said above, the task requires to compute a similarity metric from the reprensentations of the models, which implies choosing a pooling function and a layer. In Table, \ref{table:SEMANTICS}, we show a systematic exploration of three pooling functions (mean, max and min) across all the possible layers of the networks (3 layers for the LSTMs and 24 layers for ROBERTA). The first result comes from comparing the effect of unit size. The character LSTM gives very weak correlations with human scores compared to the BPE model, which itself gives weaker results than the word model. This suggests that models have a hard time extracting semantic representations when their input units do not match the linguistically relevant ones (which is in this case, the word). Yet, this is not an absolute rule since a strong BPE model like ROBERTA can outperform the word LSTM. The second result comes from comparing the correlations found at different layers. In general terms the strongest correlations are to be found in the first half of the layers. This is most apparent in ROBERTA, where the strongest score is in layer 8 out of 24. For the character LSTM and BPE it is in layer 1 out of 3, and for the word LSTM in layer 2 out of 3. Finally, the pooling method matters also, but the difference is not very large. Surprisingly, the min pooling method gives best results in 3 models.

\begin{table*}[t]
\centering
\begin{tabular}{ll @{\hspace{0.1\tabcolsep}} cccc}
\hline
\bf{model}&\bf{unit level}                             &\bf{nb units}   & \bf{Lexical}      & \bf{Syntactic}     & \bf{Semantic}\\
\hline
        &   char                                       & 30             & {\textbf{93.8}} & 63.02          &2.68        \\
        &   phone                                      & 42             & {90.92}         & 62.81          & 2.80       \\
        &   word*                                      & 40k            & {24.76}        & \bf{66.68}          &\bf{32.96}  \\
        &   word/char fallback                         & 20k            & {90.83}         & 65.78          & 20.61      \\
        &   word/phone fallback                        & 20k            & {87.8}         & 64.97          & 18.30      \\
        &   BPE word[char]                             & 20k            & {91.49}        & 65.92     & 20.42      \\
\multirow{-7}{*}{\parbox{3cm}{LSTM with gold\\boundaries}}&BPE word[phone]& 20k & {88.48}  & 66.17          & 20.52      \\
\hline
\vspace{-1em}\\
        &   char                                       & 29             &{\ul\bf{93.67}}  & 60.62          & 2.52       \\
\multirow{-2}{*}{\parbox{3cm}{LSTM without\\boundaries}}& phone & 41     & 90.47           & \bf{61.31}     & \bf{7.24}  \\
\\
\hline
        &  char                                        & 30             & 84.77           & 55.88          & 1.59     \\
        &  phone                                       & 42             & 80.6            & 56.98           & 4.80     \\
        &  charseq*                                    & 40k            & 91.22           & 62.13           & 13.08     \\
        &  phonseq*                                    & 40k            & 88.33           & 61.21          & 11.94     \\
        &  charseq/char fallback                       & 20k            & \bf{91.28}      &{\ul\bf{63.02}} &{\ul\bf{16.18}} \\
        &  phonseq/phone fallback                      & 20k            & 88.45           & 62.35          & 11.20     \\
        &  BPE charseq[char]                           & 20k            & 91.23           & 63.01          & 15.58     \\
\multirow{-8}{*}{\parbox{3cm}{LSTM with\\automatic boundaries (DP-Parse)}}& BPE phonseq[phone]& 20k & 88.14 & 62.81 & 12.09\\
\hline
        & unigram   char                              & 29              & 49.98           & 44.01          & -      \\
        & bigram    char                              & 30              & 50.14           & 48.19          & -      \\
        & unigram   phone                             & 41              & 50.22           & 43.24          & -      \\
        & bigram    phone                             & 42              & 51.04           & 48.56          & -      \\
        & unigram   word*                             & 40k             & -               & 47.96          & -      \\
\multirow{-6}{*}{\parbox{3cm}{Ngram\\baselines}} & bigram  word* & 40k  & -      & 52.05          & -      \\
\hline
\vspace{-1em}\\
\parbox{3cm}{ROBERTA large}& BPE word[char]           & 50k            & 96.03            & 82.06          & 33.16   \\
\vspace{-1em}\\
\hline
\end{tabular}
\caption{\textbf{Lexical, syntactic and semantic test scores} (higher is better) of LSTM models trained on LibriSpeech as function of availability of gold boundary and type of unit used for training. The units are characters (char), phonemes (phone), words (word), words with a fallback on characters (word/phone) or phonemes (word/char), or BPE based on words whose subword units are characters (word[char]) or phonemes (word[phone]). We also test LSTM models with automatic boundaries; here gold words are replaced by sequences of characters (charseq) or of phonemes (phonseq). Also shown unigram and bigram baseline scores and a 'topline' with a BPE pretrained model (ROBERTA). Bold indicates the best score obtained within each LSTMs's classes, underlined for the best score within the LSTMs without gold boundaries. * when the model encounters a unit not seen in training, the symbol \texttt{<UNK>} is used. }\label{tab:res}
\vspace{-0.7em}
\end{table*}

\section{The effect of word boundaries and phonetic encoding}
\vspace{-0.5em}

\label{sec:nobound}

Here, we report the result of our main experiment, in which we evaluate models trained under four versions of the training set: (1) character encoding plus boundaries (as used in the baselines reported above) (2) characters encoding without boundaries (the space character is removed) (3) phonetic encoding with boundaries (a space character is added to the output of the G2P) (4) phonetic encoding without boundaries. We add to these 4 training set two more versions, obtained by running the automatic segmentation algorithm to the (character or phoneme) corpora without boundaries. Since the automatic boundaries may not coincide with those of real words, we call the tokens isolated by this method character sequences (charseq) and phoneme sequences (phoneseq), respectively. 

The models are three-layer LSTMs. When the boundaries are not available, only character/phoneme LSTMs are used.  When the boundaries are available, we add three more models. (1) A word (or charseq, or phoneseq) model: we use the boundaries to construct a lexicon, which we cap at the 40k most frequent tokens. Each token is then one-hote encoded (including a special \texttt{<UNK>} token for all of the OOVs). (2) a word/charseq/phoneseq model with character/phoneme fallback: we use a smaller lexicon (20k) and instead of using \texttt{<UNK>}, we fall back on the character (or phoneme) level encodings. (3) a BPE model with 20k tokens. 

Table \ref{tab:res} shows the overall results on the three benchmarks. For pSIMI, we use the dev set to find the optimal combination of layers and pooling methods, and report the results on the test set. As expected, the results for the models with access to the gold boundaries are better than for the models without boundaries. The decrement in performance depends on the task (very low for pWUGGY: 2\% in relative error rate, moderate for pBLIMP: 14\%, very large for pSIMI: 28\%). Interestingly the best models depend on the tasks (with character models being best for pWUGGy, and word models for the other two). Phone models suffer from a small decrement in mosts tasks. Finally, we show that automatically generated word boundaries using unsupervised word segementation can help on two tasks (pBLIMP and pSIMI) with improved performances compared to the no-boundary condition. This is encouraging given that the automatic word boundaries are far from perfect (40\% token F-score).

\vspace{-0.7em}
\section{Conclusions}
\vspace{-0.5em}

We introduced one new dataset (pWUGGY) and two adaptations of text-based datasets (pBLIMP and pSIMI) which enable human-comparable testing of language models on character or phoneme inputs, with or without word boundaries. We show that models without word boundaries underperform models that have boundaries which can rely on higher order units like words or BPEs, and that part of this decrement can be compensated by using automatically generated word boundaries using unsupervised word segementation. We show that even relatively modest word boundary performances (40\% F-Score) can yield improvement compared to the no boundary condition. This represents the first step towards evaluating and improving language models trained from speech inputs.




\bibliography{main_bib}
\bibliographystyle{acl_natbib}

\appendix

\clearpage
\begin{center}
\textbf{\large Supplementary Materials}
\end{center}
\setcounter{equation}{0}
\setcounter{figure}{0}
\setcounter{table}{0}
\setcounter{page}{1}
\makeatletter
\renewcommand{\theequation}{S\arabic{equation}}
\renewcommand{\thefigure}{S\arabic{figure}}
\renewcommand{\thetable}{S\arabic{table}}
\renewcommand{\bibnumfmt}[1]{[S#1]}
\renewcommand{\citenumfont}[1]{S#1}

\section{Automatic segmentation with DP-Parse}\label{SM:DPPARSE}
In unsupervised word segmentation, we are given a corpus of unsegmented phonetic or character-based sentences, i.e sentences were word boundaries have been removed. The aim of this task is to find as many real word boundaries on those unsegmented sentences.

In order to tackle the task of unsupervised word segmentation, we propose a new algorithm: the Dirichlet Process Parse or DP-Parse that is inspired from the work of Goldwater et al \cite{goldwater}. 
As in \cite{goldwater} the segmentation process is composed of three steps: an initialisation, a generative unigram modelling and an inference step. The whole process is iterated until convergence.\\
First, the pipeline is initialized with a word lexicon that contains candidate word tokens along with their counts, i.e the number of times each token occurs in the corpus. At initialisation, these candidate word tokens are chosen to be all the short sentences (less than twenty phonemes long) in the corpus.\\
The iterative process starts with the generative unigram modelling. Let us consider a sentence $s$ that can be segmented in a series of $l$ candidate word tokens $s=w_1,...w_l$. The unigram model assigns a probability to $s$ as the product of the probabilities of its candidate word tokens: $P(s)=\prod_{i=0}^{l}P(w_i)$. The probability of a word token is the probability of that series of phonemes to be a real word token. It is computed using an instance of a Dirichlet Process in the way that \cite{goldwater} describes it from which we will give a general overview.

The unigram model will produce the probability of a word token $w_i$ to be a word conditioned on already segmented words $w_{-i}$. Let $i-1$ be the number of tokens already segmented, $M$ the length in phonemes of the current token $w_i$, and $n_l$ the number of other tokens already segmented that have the same label as $w_i$. $n_l$ is stored in the lexicon create at the beginning of that iteration. According to the Chinese restaurant approach to the Dirichlet Process, the probability of a token $w_i$ depends on $w_i$ being novel (i.e already in the lexicon) or not.

(1.a) P($w_i$ is novel)=$\frac{\alpha_0}{i-1+\alpha_0}$

(1.b) P($w_i$ is not novel) = $\frac{i-1}{i-1+\alpha_0}$ \\

(2.a) P($w_i=l | w_i$ is not novel) = $\frac{n_l}{i-1}$ 

(2.b) P($w_i=x_1...x_M | w_i$ is novel) = 

\quad $P_0(w_i=l)=p_\#(1-p_\#)^{M-1}\prod^{M}_{j=1}P(x_j)$ 

Giving:
\begin{equation*}
P(w_i=l |w_{-i})= \frac{n_l}{i-1+\alpha_0}+\frac{\alpha_0P_0(w_i=l)}{i-1+\alpha_0}
\end{equation*}
$\alpha_0$ is the concentration parameter that conditions the number of clusters that will be found and $P_0$ is the base distribution that determines the characteristic of each word clusters.\\
\\
Now that each candidate word tokens can be assigned a probability, word boundaries can be drawn during the inference step. We replaced the Gibbs sampler used by \cite{goldwater} by a dynamic programming inference in order to speed up the segmentation process. In the Gibbs sampling approach, each boundary of the sentence $s$ is sampled one at a time conditioned on all previous boundaries. In comparison, our method samples all boundaries in $s$ at once. It works as if it samples a parse of the sentence $s$, hence the name of the pipeline: DP-Parse.
In order to explain our sampling process, let us suppose we have $s$ possible parses for a phonetic sentence $P$, such that $s=(s_1,...,s_P)$. These parses can be represented in a table that contains the costs of boundaries between all possible pairs of word tokens. A parse is a path in that table that covers the whole sentence. Then to find the cost of each parse, we perform a dynamic programming beam search that returns the N parses that have the lowest log probabilities. The segmentation of $s$ will result from sampling one parse randomly among those N best parses. We do not choose the best parse for $s$ in order to avoid being stuck in poor local optima. Once a parse is sampled for $s$ , the next sentence is considered and so on until the end of the corpus. \\
Finally, the process can iterate. A new lexicon is populated with found word tokens, new probabilities can be computed for each word tokens according to the unigram generative model. Iterations last until the probability of the whole corpus’ segmentation does not decrease anymore. \\ 

Regarding performances, DP-Parse works slightly sub-optimally compared to a unigram model trained in the Adaptor Grammar framework (AG) \cite{goldwater}, but runs in average 50 times faster. On a 5\% subset of the Librispeech-960, AG reaches a token F1 score of $0.64$ where our DP-Parse gets $0.57$. However, regarding time efficiency, running 10 iterations of DP-Parse on the full Librispeech-960 takes 5 hours whereas AG would need around 10 days. 

\section{Sampling method to balance ngram scores}\label{SM:sampling}
We describe here our sampling method to balance ngram scores for pWUGGY and pBLIMP datasets.
We first show the algorithm that we applied to pWUGGY, then we just modify slightly the algorithm for the pBLIMP dataset.

For WUGGY, let's assume that we have $N$ words $w_1,\dots, w_N$; and for each word $w_i$, we have a list of $K$ matching nonword candidates $nw_i^1,\dots, nw_i^K$. We also assume that each word or nonword $w$ has $M$ scores $s_1(w),\dots,s_M(w)$ (this might be unigram/bigram char/phone scores). We aim to choose, for each word $w_i$, a matching nonword $nw_i^*$ such that the proportion of the pairs where the score of the word is higher than the score of nonword is close to 50\% as possible, for each of $M$ scores.

In other words, we want to build a list of word-nonword pairs $L=\{(w_1, nw_1^*),\dots,(w_N, nw_N^*)\}$ such that the objective function
\begin{equation}\label{eq:scoreObj}
    \text{obj}(L) = \sum_{m=1}^M\lvert\text{accuracy\_of\_score\_m(L)-0.5}\rvert
\end{equation}
is as close to zero as possible.

We thus deduce a simple sampling method as follows: We first initialize a list $L$ of chosen pairs of word and nonword. At each iteration, we randomly choose an unchosen word. Then we sample a nonword candidate in the list of matching nonword candidates, update the list with the new pair, and compute the objective function of the new list as given in \ref{eq:scoreObj}. If the objective increases, we remove this newly added element, and resample a new nonword from the list of candidates. If we encounter all the nonword candidates but cannot find a new pair, we random choose a nonword from the list of candidates. We then continue to the next word until all the words are chosen.

We found afterwards that if we sample all the words at the same time, we can obtain an overall score very close to 50\%, but then words with high frequency or with short length tended to have higher accuracy than others. We then decided to divide the words into sub-categories by frequency and word length, and then do the sampling on each of the sub-categories, which gives a more balanced score on all the length and frequency levels.

For BLIMP, the candidates are slightly different. We now have a list of $N$ pairs of grammatical and non-grammatical sentences and we want to choose $K$ pairs among them such that the accuracy of the chosen pairs is as close to 50\% as possible as for WUGGY.
We can then use the same sampling method as described above, with the exception that instead of choosing a word and sampling the nonword candidates at each iteration, we sample an unchosen pair in the list of candidates, and add that pair to the chosen list if we succeed to decrease the objective function.

As we also found that there is a huge difference in the accuracy scores of linguistic paradigms, we tried to do the sampling by each sub-paradigm. However, there were still some paradigms that we were not able to perfectly balance the score.

\end{document}